\documentclass{article} 
\usepackage{nips13submit_e,times}
\usepackage{hyperref}
\usepackage{url}
\usepackage{amsmath}
\usepackage{multirow}
\usepackage[sort]{natbib}

\title{Deep Learning Embeddings for Discontinuous Linguistic
  Units}

\author{
Wenpeng Yin and Hinrich Sch\"{u}tze\\
Center for Information and Language Processing\\
University of Munich\\
Germany\\
\texttt{wenpeng@cis.lmu.de}
}

\nipsfinalcopy 

\def\secref#1{Section~\ref{sec:#1}}
\def\seclabel#1{\label{sec:#1}\label{p:#1}}
\def\eqref#1{Eq.~\ref{eqn:#1}}

\newcommand{\enoteson}{\long\gdef\enote##1##2{{\large\bf
\hspace{1cm}$<<<$ ##1: ##2 $>>>$\hspace{1cm}}}}
\enoteson

\begin{document}

\maketitle

\begin{abstract}
Deep learning embeddings have been successfully used for many
natural language processing problems. Embeddings are mostly
computed for word forms although a number of recent papers
have extended this to other linguistic units like morphemes
and phrases. In this paper, we argue that learning
embeddings for \emph{discontinuous linguistic units} 
should also be considered. 
In an experimental evaluation on
coreference resolution, we show that such embeddings perform
better than word form embeddings.
\end{abstract}

\section{Motivation}
One advantage of recent work in deep learning on natural
language processing (NLP) is that linguistic units are represented
by rich and informative embeddings. These embeddings  support better
performance on a variety of NLP tasks \citep{collobert2011natural} than symbolic linguistic
representations  that do not directly
represent information about similarity and other linguistic
properties. Embeddings are mostly derived for word forms
although a number of recent papers have extended this to
other linguistic units like morphemes \citep{luong2013better} and phrases \citep{mikolov2013distributed}. 
Thus, an important question is: what are the basic
linguistic units that should be represented by embeddings in
a deep learning NLP system? In this paper, we argue that
certain discontinuous linguistic units should also have
embeddings. 
We will restrict ourselves to the arguably simplest possible
type of discontinuity: two noncontinous words. For example,
in the sentence ``this tea helped me to relax'',
``helped*to'' is one of several such two-word discontinuities.
We will refer to discontinuous linguistic units like ``helped*to''
as \emph{minimal contexts} (MC) for reasons that will become
clear presently.

We can approach the question of what basic linguistic units
should have representations from a practical as well as
from a cognitive point of view. In practical terms, we want
representations to be optimized for good
generalization. There are many situations where 
a particular task involving a phrase cannot be solved based
on \emph{the phrase itself}, but it can be solved by analyzing \emph{the
  context of the phrase}. For example, if a coreference
resolution system needs to determine whether the unknown word
``Xiulan'' (a Chinese first name) in ``he helped Xiulan to
find  a flat'' refers to an animate or an
inanimate entity,
then the minimal context
``helped*to'' is a good indicator for the animacy of the
unknown word -- whereas the unknown word itself provides no
clue.

From a cognitive point of view, it can be argued that many
basic units that the
human cognitive system uses are also
discontinuous. Particularly convincing examples for such
units are phrasal verbs in English, which frequently occur
discontinuously. It is implausible to suppose that we
retrieve atomic representations for, say, ``keep'', ``up'',
``under'' and ``in'' and then combine them to form the meanings of phrases like ``keep him
up'', ``keep them under'', ``keep it in''. Rather, it is
more plausible that we recognize ``keep up'', ``keep under''
and ``keep in'' as relevant basic linguistic units in these
contexts and that the human cognitive systems represents
them as units.

This paper presents an initial study of minimal context embeddings and
shows that they are better suited for a classification task
needed for coreference resolution than word embeddings.
Our conclusion is that minimal contexts (as well as
inflected word forms, morphemes and phrases) should be
considered as basic units that we need to learn embeddings for.

\section{Experimental setup}\seclabel{setup}
\subsection{Embedding learning}
\seclabel{embedding}
With English Gigaword Corpus, we use the \emph{skip-gram model} as implemented in
word2vec\footnote{\url{https://code.google.com/p/word2vec/}}
\citep{mikolov2013distributed} to induce
embeddings.  To be able to use word2vec directly without
code changes, we represent the
corpus
as a sequence of sentences, each
consisting of two tokens: an MC (written as the two enclosing words
separated by a star) and a word that occurs between
the two enclosing words. The distance $k$ between the two
enclosing words can be varied. In our experiments, we 
use either distance $k=2$ or distance $2
\leq k\leq 3$. For
example, for $k=2$,
the trigram 
$w_{i-1}~w_{i}~w_{i+1}$ generates the single sentence
``$w_{i-1}\mbox{*}w_{i+1}~w_{i}$''; and for $2\leq k\leq 3$, the fourgram
$w_{i-2}~w_{i-1}~w_{i}~w_{i+1}$ generates the four sentences
``$w_{i-2}\mbox{*}w_{i}~w_{i-1}$'',
``$w_{i-1}\mbox{*}w_{i+1}~w_{i}$'',
``$w_{i-2}\mbox{*}w_{i+1}~w_{i-1}$'' and
``$w_{i-2}\mbox{*}w_{i+1}~w_{i}$''.


Note that the reformated corpus enables word2vec to learn embeddings for single
words and MCs simultaneously, we discard
the word embeddings, and yet compute standard word embeddings on the original corpus using
word2vec skip-gram model. In experiments, embedding size is set to 200.

\subsection{Markable classification task}
A \emph{markable} is a linguistic expression that refers to an
entity in the real world or another linguistic
expression. Examples of markables include
noun phrases (``the man''), named entities (``Peter'')
and nested noun phrases (``their'').
We address the task of \emph{animacy classification} of
markables: classifying them as
animate/inanimate. This feature is useful for coreference
resolution systems because only animate markables can be
referred to using masculine and feminine pronouns in English
like ``him'' and ``she''. Thus, this is an important clue
for automatically clustering the markables of a document into correct
coreference chains.

To create training and test sets, we extract
all 39,689 coreference chains from the CoNLL2012
OntoNotes corpus.\footnote{\url{http://conll.cemantix.org/2012/data.html}}
We label chains that contain one of the markables ``she'', ``her'', ``he'',
``him'' or ``his'' as animate and chains that contain
one of   ``it'' or ``its'' as inanimate.

We extract 
39,942 markables and their corresponding MCs 
from the 10,361 animate and inanimate chains
where an MC simply is the pair of
the two words occurring to the left and right of the
markable.  The gold label of a markable and its MC is the animacy
status of its chain: either animate or inanimate.  We divide
all MCs having received an embedding in the
embedding learning phase into 
a training set of 11,301 (8097 animate, 
3204 inanimate) and a balanced test set
of 4036.

We use 
LIBLINEAR\footnote{\url{https://github.com/bwaldvogel/liblinear-java}}
for classification, with penalty factors 3 and 1 for
inanimate and animate classes, respectively, because the
training data are unbalanced.

\section{Experimental results}\seclabel{results}
We compare the following representations for animacy classification
of markables. (i) MC: minimal context embeddings with $k=2$
and $2 \leq k \leq 3$; (ii) concatenation:
concatenation of the embeddings of the
two enclosing words where the embeddings are either standard
word2vec embeddings (see \secref{embedding}) or the embeddings
published
by \citet{collobert2011natural};\footnote{\url{http://metaoptimize.com/projects/wordreprs/}}
 (iii) the bag-of-words (BOW) 
representation of a minimal context:
the concatentation of two one-hot vectors of 
dimensionality $V$ where $V$ is the size of the
vocabulary. The first (resp.\ second) vector is the one-hot vector for the
left (resp.\ right) word of the MC. Experimental results are shown in Table \ref{tab:result}.

\begin{table}[ht]
\centering
\begin{tabular}{l|l|r@{}l@{}l}
\multicolumn{2}{c|}{representation}&
\multicolumn{3}{l}{accuracy}\\\hline\hline
\multirow{2}{*}{MC} &$k=2$ &0.703& \\
&$2\leq k \leq 3$  &  0.700& \\\hline
\multirow{2}{*}{concatenation} &skip-gram model &0.668&*&$^{\dagger}$\\
& C\&W  & 0.662&*&$^{\dagger}$ \\\hline
BOW& & 0.638&*&$^{\dagger}$
\end{tabular}
\caption{Classification accuracy. Mark ``*''  means
  significantly lower than MC, $k=2$;  ``$\dagger$'' means
  significantly lower than MC, $2 \leq k \leq 3$.}\label{tab:result} 
\end{table}

The results show that MC
embeddings have an obvious advantage in this classification task,
both for $k=2$ and $2\leq k \leq 3$. This validates our hypothesis
that learning embeddings for discontinuous linguistic units
is promising.

In our error analysis, we found two types of frequent
errors.  (i) {\bf Unspecific MCs.} Many MCs are equally
appropriate for animate and inanimate markables. Examples of
such MCs include ``take*in'', ``keep*alive'' and
``then*goes''.  (ii) {\bf Untypical use of specific MCs.}
Even MCs that are specific with respect to what type of
markable they enclose sometimes occur with the ``wrong''
type of markable.  For example, most markables occurring in the MC ``of*whose'' 
are animate because ``whose'' usually refers to
an animate markable. However,
in the context
``\ldots the southeastern area of Fujian whose
economy is the most active'' the enclosed markable is
Fujian, a province of China. 
This example shows that ``whose'' occasionally refers to an
inanimate entity
even though these cases are
infrequent.


\section{Related work}
Most work on embeddings has focused on word forms with a few
exceptions, notably embeddings for stems and morphemes
\citep{luong2013better} and for phrases
\citep{mikolov2013distributed}.  To the best of our
knowledge, our work is the first to learn embeddings for
discontinuous linguistic units.

An alternative to learning an embedding for a linguistic
unit is to calculate its distributed representation from the
distributed representations of its parts; the best known
work along those lines is 
\citep{socher2010learning,socher2011semi,socher2012semantic}. 
This approach is superior for units that are compositional,
i.e., whose properties are systematically predictable from
their parts. Our approach (as well as similar work on
continuous phrases) only makes sense for noncompositional
units.

\section{Conclusion and Future Work}
We have argued that discontinuous linguistic units are part
of the inventory of linguistic units that we should compute
embeddings for and we have shown that such embeddings are
superior to word form embeddings in a coreference resolution
task.

It is obvious that we cannot and do not want to compute
embeddings for all possible discontinuous linguistic
units. Similarly, the subset of phrases that embeddings are
computed for should be carefully selected. In future work,
we plan to address the question of how to select a subset of
linguistic units -- e.g., those that are least compositional
-- when inducing embeddings.

\bibliographystyle{elsarticle-harv}
\bibliography{ICLR}

\end{document}